\documentclass[runningheads]{llncs}
\usepackage[T1]{fontenc}
\usepackage{graphicx}
\usepackage[utf8]{inputenc}
\usepackage[T1]{fontenc}
\usepackage{multirow}
\usepackage{subcaption}

\usepackage[table,xcdraw]{xcolor}
\usepackage{amsmath}
\begin{document}
\title{Enabling Tactile Feedback for Robotic Strawberry Handling using AST Skin}
\titlerunning{Acoustic Soft Tactile Skin}
%
\author{Vishnu Rajendran S\inst{1} \and
Kiyanoush Nazari\inst{2} \and
Simon Parsons\inst{1} \and
Amir Ghalamzan E\inst{3}}
\authorrunning{V. Rajendran et al.}
%
\institute{Lincoln Institute of Agri-food Technology, University of Lincoln, UK, 
\email{25451641@students.lincoln.ac.uk}\\ 
\and
School of Computer Science, University of Lincoln, UK, 
\email{sparsons@lincoln.ac.uk}\\
\and
University of Surrey, UK, \email{a.esfahani@surrey.ac.uk}\\
}
\maketitle              
\begin{abstract}
Acoustic Soft Tactile (AST) skin is a novel sensing technology which derives tactile information from the modulation of acoustic waves travelling through the skin's embedded acoustic channels. A generalisable data-driven calibration model maps the acoustic modulations to the corresponding tactile information in the form of contact forces with their contact locations and contact geometries. AST skin technology has been highlighted for its easy customisation. As a case study, this paper discusses the possibility of using AST skin on a custom-built robotic end effector finger for strawberry handling. The paper delves into the design, prototyping, and calibration method to sensorise the end effector finger with AST skin. A real-time force-controlled gripping experiment is conducted with the sensorised finger to handle strawberries by their peduncle. The finger could successfully grip the strawberry peduncle by maintaining a preset force of 2 N with a maximum Mean Absolute Error (MAE) of 0.31 N over multiple peduncle diameters and strawberry weight classes. Moreover, this study sets confidence in the usability of AST skin in generating real-time tactile feedback for robot manipulation tasks.

\keywords{Soft Tactile skin \and Acoustics  \and Grip-force control.}
\end{abstract}
\section{Introduction}

\begin{figure*}[tb!]
      \centering
        \includegraphics[width=1\textwidth]{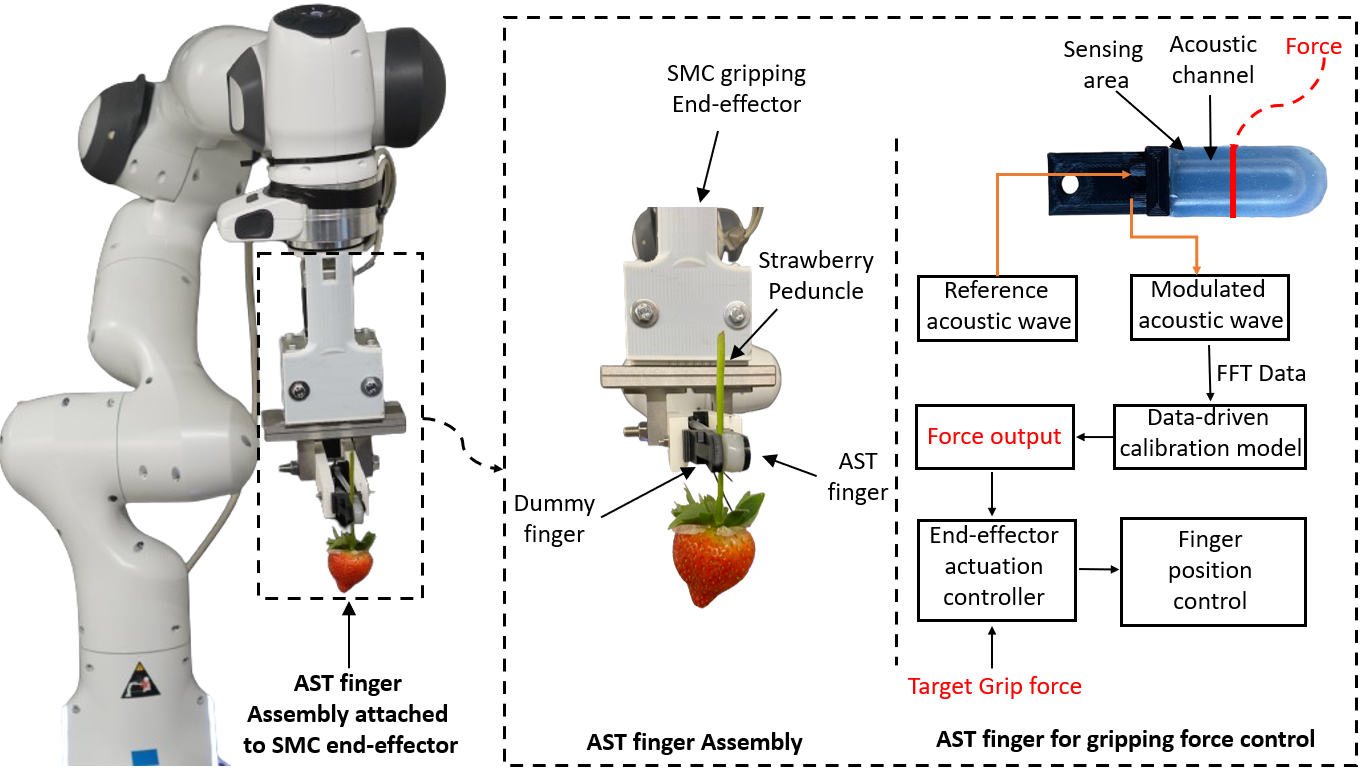}
\caption{Demonstration of AST sensing technology to generate tactile force feedback for a robot manipulation task: The AST Skin is attached to a custom end effector finger (the AST finger), and the force feedback from the AST finger is used in performing force-controlled handling of strawberries by gripping their peduncle. This finger assembly is mounted on an SMC gripper for validation purposes, but later, it will be attached to a custom-built strawberry harvesting end-effector}
\label{overview}
\end{figure*}

Soft tactile sensors are generally used in manipulation tasks that involve handling soft deformable objects. These sensors are typically characterised by soft deformable skin whose deformation is converted to tactile feedback as tactile readings using an integrated transduction mechanism. There has been continuous progress in the development of soft tactile sensors using various skin materials and transduction mechanisms~\cite{roberts2021soft,wang2023tactile}.

Various robotic applications call for customisable soft tactile sensors. This is especially important to generate tactile feedback from robot counterparts for the effective execution of manipulation tasks. The customisation can be in the form of sensor shape and size, material and sensing specifications (such as sensitivity, measurement range) to fit the requirement. Since the conventional soft-tactile sensors have their transduction mechanism integrated with the sensing skin, customisation is not straightforward. Usually, the transduction mechanism uses electric principles (e.g., resistive~\cite{zimmer2019predicting}, capacitive~\cite{li2016wide}, piezoelectric~\cite{song2019pneumatic}, magnetic~\cite{rehan2022soft,diguet2022tactile}, impedance~\cite{wu2022new}) non-electric principles (e.g., camera-based~\cite{ward2018tactip,gomes2022geltip,lambeta2020digit,10017344,sferrazza2019design}, fluid-based~\cite{gong2017pneumatic}) or their combinations~\cite{park2022biomimetic}. When it comes to sensors using electric principles, they have closely knit circuitry embedded beneath the sensing surface and it is the camera with its accessories for camera-based sensors. Similarly, fluid pressure-based sensors have intricate fluid lines integral to the sensing skin. Hence, customising the skin requires considerable effort to also customise the integrated transduction mechanism elements. However, the possibility of keeping the sensing skin and transduction elements modular offers a better scope for customisation. This is possible when the transduction mechanism uses propagating mediums (such as vision/light, fluids or acoustics) to sense the disturbances on the skin by analysing their respective modulations. While using vision, the camera and its accessories result in a bulky form factor of the sensor~\cite{wei2015overview}. Hence, using a camera provides limitations in minimising the form factor of the whole sensor. Moreover, fluid-based methods are known for their delayed measurement responses~\cite{fujiwara2023agar}. In this context, acoustics has a promising scope. It only requires minimal hardware components, typically a speaker and a microphone~\cite{ono2015sensing,wall2023passive}. Most importantly, it can potentially derive diverse tactile information such as force, contact location, temperature and contact material nature~\cite{wall2023passive}.

 With the modularity concept for customisation and harnessing the capabilities of acoustics, a low-cost novel tactile sensing technology, namely, Acoustic Soft Tactile skin (AST Skin), has been developed~\cite{rajendran2023acoustic}. This sensing technology keeps the soft sensing skin and transduction mechanism elements modular. The sensing skin only needs hollow acoustic channels beneath the sensing surface. A speaker and microphone unit form the transduction elements that need to be connected to the acoustic channel. A reference acoustic wave emitted by a speaker propagates through these channels and returns to the microphone. When external forces act on the sensing surface, they deform the channel; hence, the modulation of the acoustic waves varies. This variation is used to measure tactile interactions. Moreover, AST skin technology uses a generalisable data-driven machine-learning model for its calibration. Such a calibration model can account for the variation of the sensing skin's form factor or material change that may arise from customisation, which is complex to establish through an analytical model. AST skin technology has been proven to measure normal forces, 2D contact locations, and contact surface geometries~\cite{rajendran2023acoustic,10522056}. Moreover, the skin's resilience to external sound disturbances is also validated. 

In previous work, various skin configurations and their impact on tactile measurements are studied in detail~\cite{rajendran2023acoustic}. This paper evaluates the customisability of AST skin by integrating it on a custom end effector finger (the AST finger) and tests its sensing capabilities (refer to figure~\ref{overview}). This finger has been developed as a retrofit to a strawberry harvesting robotic end-effector. The AST finger is expected to grip the strawberry peduncle and provide tactile feedback about the peduncle grip status when the robot's vision unit cannot confirm the grip due to occlusions. Moreover, the feedback from the AST finger facilitates effective grip force control for the strawberry handling involved in the harvesting cycle. This paper does not discuss the details of the harvesting end effector mechanism; instead, the AST finger is tested using a general purpose two jaw gripping end effector. The proposed AST finger has a reduced size factor compared to other sensorised gripper fingers available for strawberry handling~\cite{visentin2023soft}. The size of the fingers impacts its usability in handling strawberries grown in clusters. Moreover, gripping the strawberry body while harvesting can lead to fruit bruising~\cite{aliasgarian2013mechanical}, reducing its shelf life. Thus, the AST finger aims to grip the peduncle rather than the fruit body. 

The remainder of the paper outlines the methodology to develop the AST finger and its sensing performance during the real-time strawberry peduncle gripping trials.

\section{Methods}

This section discusses the methodology for the end-to-end development of the AST finger. Initially, it outlines a study to determine the maximum gripping force that can be applied to the strawberry peduncle. This maximum grip force value helps to decide the force range for which the AST skin needs to be calibrated. Later in this section, the design of the AST finger, prototyping details, and the adopted data-driven calibration approach are detailed.

\subsection{Study on Gripping Force} 

During harvesting, the robot's end effector fingers should grip the peduncle before the cutting and continue holding it until the detached strawberry is placed in the onboard storage (e.g., punnet). Applying the proper force is essential to prevent the strawberry from slipping off the fingers during this manipulation cycle. There are two reasons for slippage: applying insufficient force or crushing the peduncle due to excessive force. 

To determine the safe range of force that can be applied to the peduncle, a compression study was conducted with peduncle samples from two breeds of strawberries~\cite{10053882}. The study found that the safe limit of gripping force can be approximated to 10 N. Hence, the AST finger is calibrated for a range of 0 to 10 N. 

\subsection{AST Finger Design}

\begin{figure}[tb!]   
\centering
    \begin{subfigure}[t]{0.4\textwidth}  
    \includegraphics[width=\textwidth]{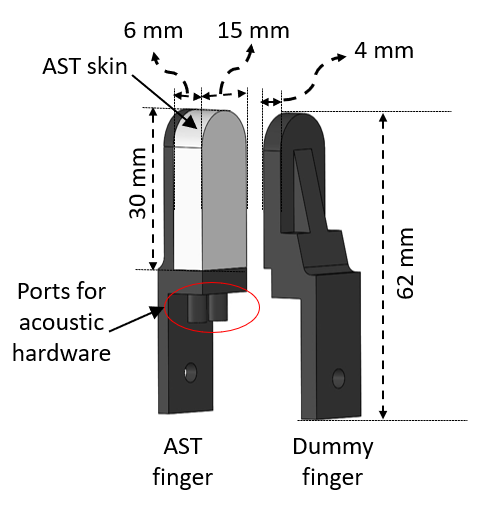}
    \caption{} \label{2a}
\end{subfigure}
\begin{subfigure}[t]{0.4\textwidth} 
    \includegraphics[width=\textwidth]{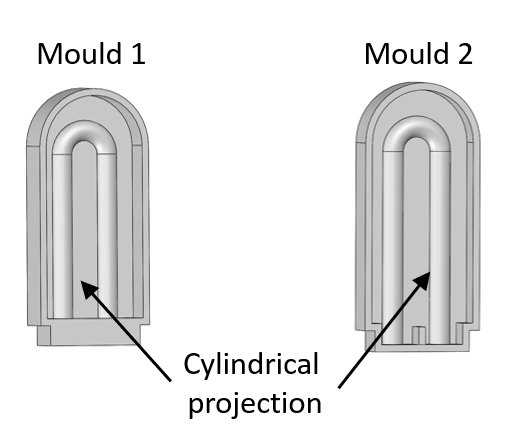}
    \caption{} \label{2b}
\end{subfigure}\hfill
\caption{(a). AST finger with its mating dummy finger, (b). Skin mould: The two semi-cylindrical projections in each mould help to create the skin's cylindrical cavity while joining two halves of the cured silicone}
\label{design}
\end{figure}

The AST finger comprises of a back plate on which the AST skin is attached (refer to figure~\ref{2a}). The back plate is 3D printed using PLA material, while the AST skin is made by moulding. The AST skin is moulded in two halves separately and then joined together. Figure~\ref{2b} shows the mould design. The two moulds are 3D printed, and a Silicone rubber compound with a 10 A shore hardness value (PlatSil Gel 10) is poured into the moulds. After the curing period, the two halves of skin are joined together to form the whole skin. Later, the AST skin is attached to the back plate. The same silicone material is used as the adhesive to join the skin halves and to attach it to the back plate. The acoustic channel is given a 3 mm diameter and is at 1 mm below the sensing surface of the skin so that a gentle touch can affect the channel geometry. 
The AST finger has two ports to connect the modular acoustic hardware components (refer to figure ~\ref{2a}). In this prototype, a regular headphone speaker and microphone are connected to the port via a flexible tube as a test case. In the future, a miniature speaker-microphone will be used.

As a mating finger for the AST finger to grip the strawberry peduncle, a dummy finger has been 3D printed with PLA material. To actuate both fingers for the gripping trials, they are attached to an SMC gripper as shown in figure~\ref{overview} and~\ref{4a}.

\subsection{Skin Calibration}

\begin{figure}[tb!]
      \centering
        \includegraphics[width=0.5\textwidth]{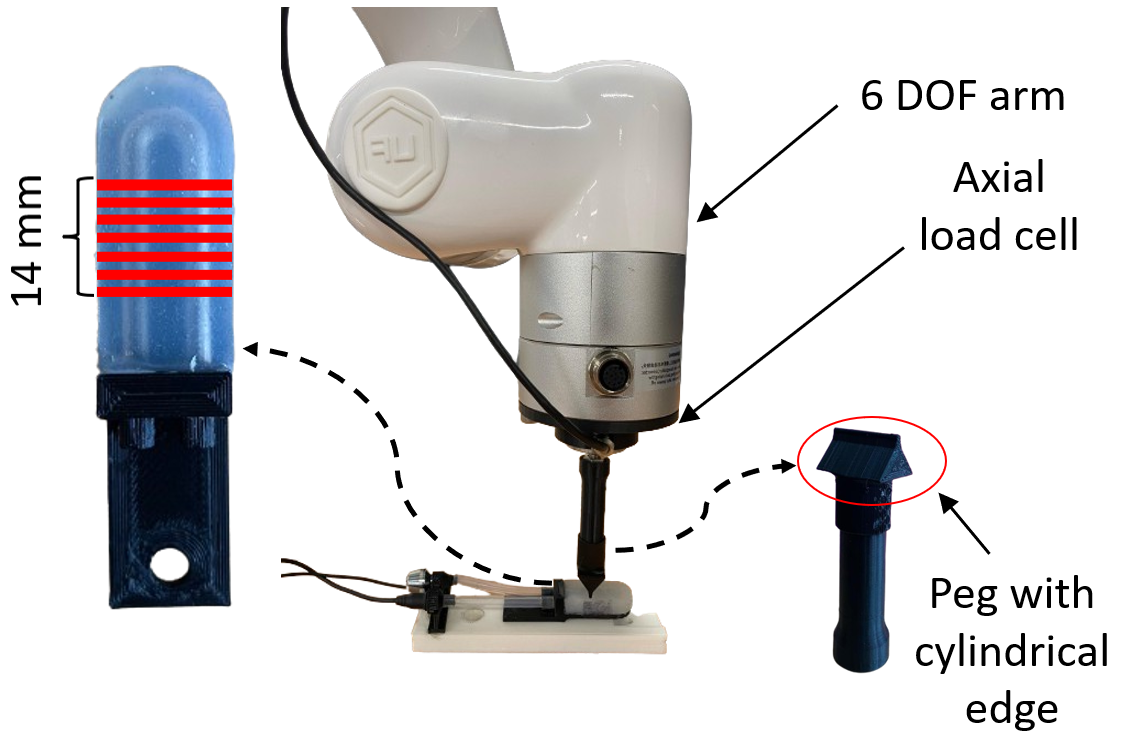}
\caption{AST finger calibration: The calibrated area of the AST finger (left), the calibration set up consisting of 6 DOF UFactory xArm, axial load cell with peg attached to the robot arm wrist (middle), the calibration peg with cylindrical profiled edge to simulate the shape of strawberry peduncle (right)}
\label{cal}
\end{figure}

As mentioned in Section 1, the AST skin uses a data-driven calibration model. In here, the AST finger is only calibrated to measure contact forces. So, the data set used for the calibration only involves the Fast Fourier Transform (FFT) of the modulated acoustic wave and the corresponding force that causes the modulation. To derive this dataset, a robot-based calibration setup is used to apply known forces on the sensing surface, and the corresponding FFT of the acoustic wave modulation is recorded against the applied force. The robot wrist is fitted with an axial load cell-peg assembly, and the peg has a cylindrical profile with a diameter of 1.5 mm (refer to the figure~\ref{cal}, right). This cylindrical profile simulates the average diameter of the strawberry peduncle considered for studying the gripping force limit. During actual deployment on the strawberry harvesting end effector, the finger will be calibrated with a diverse data set generated with different diameter pegs covering the possible peduncle diameters and orientations ranges. 

As a test case, only a 14 mm section of the skin is calibrated, and this section was divided into seven subsections at a 2 mm gap, as shown in the figure~\ref{cal} (left). The robot drives the peg vertically downwards at subsections with an equal increment of 0.5 mm from the skin surface until the force value reaches 10 N. During this, the modulated FFT data is recorded against the force values. The resulting data set is used to train the regression machine learning model to predict force from FFT data. To select the best-suited regression model, MATLAB regression learner is used. Different regression models are trained with a data partition of 90:10 ratio and 10\% cross-validation folds (refer to table~\ref{reg}). The Exponential Gaussian Process regression model is selected based on its lower validation error (RMSE: 0.27) and used as the calibration model for the skin. 

This calibration model is tested with the 10\% test data, and the predictions obtained are presented in the table~\ref{calres}. About 91.1\% of predictions are made within $\pm0.5$ N tolerances, while it could make 99 \% of predictions at $\pm1.0  $ N tolerances. Moreover, the mean absolute error of the predictions is 0.16 N with a standard deviation of 0.22 N.

\begin{table}[]
\centering
\small\addtolength{\tabcolsep}{-1pt}
\caption{Comparison of Regression models using MATLAB Regression Learner}
\label{reg}
\begin{tabular}{|l|c|}
\hline
\textbf{Regression Models}                                         & \textbf{Validation Error (RMSE)} \\ \hline
\textbf{Linear Regression}                                &                         \\
\cellcolor[HTML]{FFFFFF}\textit{Linear}                   & 0.91                       \\
\cellcolor[HTML]{FFFFFF}\textit{Interactions Linear}      & 0.56                       \\
\cellcolor[HTML]{FFFFFF}\textit{Robust}                   & 0.97                       \\
\cellcolor[HTML]{FFFFFF}\textit{Step-wise Linear}          & 0.56                       \\ \hline
\cellcolor[HTML]{FFFFFF}\textbf{Regression Trees}         &                         \\
\cellcolor[HTML]{FFFFFF}\textit{Fine Tree}                & 0.42                       \\
\cellcolor[HTML]{FFFFFF}\textit{Medium Tree}              & 0.43                       \\
\cellcolor[HTML]{FFFFFF}\textit{Coarse Tree}              & 0.46                       \\ \hline
\cellcolor[HTML]{FFFFFF}\textbf{Support Vector Machines}  &                         \\
\cellcolor[HTML]{FFFFFF}\textit{Linear}                   & 0.95                       \\
\cellcolor[HTML]{FFFFFF}\textit{Quadratic}                & 12                      \\
\cellcolor[HTML]{FFFFFF}\textit{Cubic}                    & 101.93                      \\
\cellcolor[HTML]{FFFFFF}\textit{Fine Gaussian}            & 0.43                      \\
\cellcolor[HTML]{FFFFFF}\textit{Medium Gaussian}          & 0.49                      \\
\cellcolor[HTML]{FFFFFF}\textit{Coarse Gaussian}          & 0.59                     \\ \hline
\cellcolor[HTML]{FFFFFF}\textbf{Gaussian Process}         &                         \\
\cellcolor[HTML]{FFFFFF}\textit{Rational Quadratic}       & 0.28                      \\
\cellcolor[HTML]{FFFFFF}\textit{Squared Exponential}      & 0.40                      \\
\cellcolor[HTML]{FFFFFF}\textit{Matern 5/2}               & 0.34                      \\
\cellcolor[HTML]{FFFFFF}\textit{Exponential}              & \textbf{0.27}                      \\ \hline
\cellcolor[HTML]{FFFFFF}\textbf{Ensemble of Trees}        &                         \\
\cellcolor[HTML]{FFFFFF}\textit{Boosted Trees}            & 0.50                      \\
\cellcolor[HTML]{FFFFFF}\textit{Bagged Trees}             & 0.37                      \\ \hline
\cellcolor[HTML]{FFFFFF}\textbf{Neural Networks}          &                         \\
\cellcolor[HTML]{FFFFFF}\textit{Narrow Neural}            & 0.46                      \\
\cellcolor[HTML]{FFFFFF}\textit{Medium Neural}            & 0.43                      \\
\cellcolor[HTML]{FFFFFF}\textit{Wide Neural Network}      & 0.41                      \\
\cellcolor[HTML]{FFFFFF}\textit{Bi-layered Neural Network} & 0.44                      \\
\cellcolor[HTML]{FFFFFF}\textit{Tri-layered Neural Network} & 0.42                      \\ \hline
\end{tabular}
\end{table}

\begin{table}[]
\centering
\caption{Prediction performance of the calibration model}
\label{calres}
\begin{tabular}{|c|c|}
\hline
\begin{tabular}[c]{@{}c@{}}\textbf{Absolute} \\ \textbf{Error (N)}\end{tabular} & \begin{tabular}[c]{@{}c@{}}\textbf{Percentage} \\ \textbf{Predictions} (\%)\end{tabular} \\ \hline
$\pm0.5$                                                           & 91.13924                                                               \\ \hline
$\pm1.0  $                                                           & 99.27667                                                               \\ \hline
$\pm1.5$                                                           & 99.81917                                                               \\ \hline
$\pm2.0 $                                                            & 100                                                                    \\ \hline
\end{tabular}
\end{table}

\section{Performance Evaluation of AST Finger}

This section evaluates the usability of the AST finger's feedback for force-controlled gripping of the strawberry peduncle. 

For this, a series of trials are conducted where the AST finger grips strawberries by their peduncles for a simple pick-and-drop sequence. The AST and dummy finger are connected to an SMC LEZH gripping end effector using an extension adaptor, as depicted in figure~\ref{4a}. The end effector with the fingers is then attached to a Franka Emika robotic arm. This SMC gripper increases the gripping force by closing the fingers in adjustable grip widths.

\begin{figure}[tb!]
      \centering
        \includegraphics[width=0.6\textwidth]{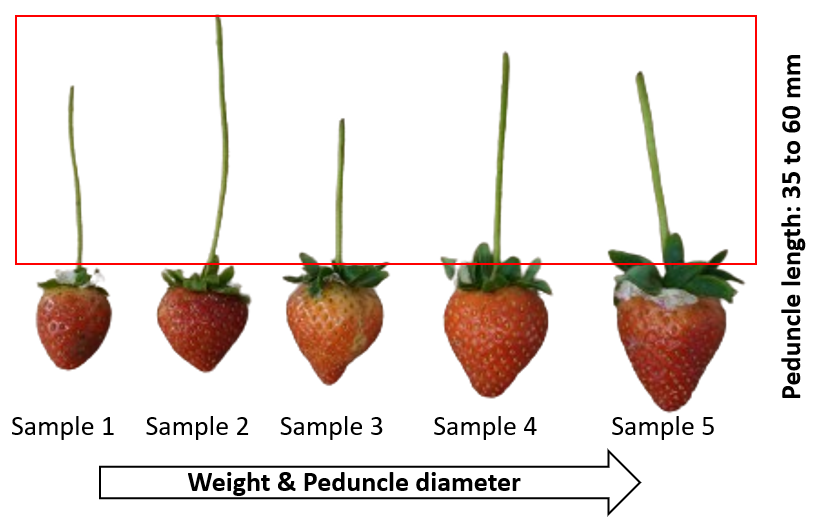}
\caption{Strawberry samples selected for the gripping trials}
\label{straw}
\end{figure}

For the gripping trials, five strawberries are collected with 35-60 mm peduncle left on them (refer to figure~\ref{straw}). These strawberries have an average weight and peduncle diameter of 0.141 N and 1.73 mm, respectively (refer to table~\ref{strspec}). A gripping force of 4 N is assumed for gripping these strawberries during the pick-and-drop manipulation cycle. With this gripping force, a mass of 91 grams (approximately 1 N) can be manipulated when the manipulator acceleration is kept below 1.0 m/s$^2$ and when the gripping surface of the finger has a friction coefficient of 0.5 (refer to the equation~\ref{eqn1} and with S as 2). Generally, the friction coefficient of silicone material is closer to 1, but it is assumed to be 0.5 here, providing an additional safety factor. 
Since the AST finger assembly forms a 2-jaw finger configuration, the total gripping force of 4 N can be considered shared between the two fingers. Hence, 2 N is the target force to be read from the AST finger to complete the gripping action before the pick-and-drop traverse. 

\begin{equation}
F_g=\frac{m.(g+a).S}{\mu}
\label{eqn1}
\end{equation}
where; 'F$_g$' is the net gripping force (N), 'm' is the mass to be handled (Kg), 'g' is the acceleration due to gravity (m/s$^2$), '$\mu$' is the coefficient of friction, and 'S' is factor of safety~\cite{Grippingforce}.

\begin{table}[]
\centering
\caption{Features of strawberries used for the gripping trials}
\label{strspec}
\begin{tabular}{|c|c|c|}
\hline
\textbf{Sample}  & \textbf{Weight (N)}   & \textbf{Peduncle Diameter (mm)}   \\ \hline
1      & 0.084   & 1.24 \\ \hline
2      & 0.111   & 1.38 \\ \hline
3      & 0.155   & 1.88 \\ \hline
4      & 0.176   & 1.90 \\ \hline
5      & 0.181   & 2.29 \\ \hline
Mean   & 0.141   & 1.73 \\ \hline
\end{tabular}
\end{table}

\subsection{Gripper Force Controller}
Tactile feedback-based reactive grip force control is commonly used for grip stabilisation in robotic manipulation~\cite{deng2020grasping}. Here, the real-time force feedback ($f_m$) from the AST finger is leveraged for such a grip force controller. As such, the controller is set to stop gripping by settling the measured grip force ($f_m$) within a safety zone close to the desired grip force $[f_d - \epsilon, f_d + \epsilon]$~(refer to equation~\ref{gripeq}). As described earlier, the desired grip force ($f_d$) is 2 N. For this control implementation, the grip width of the SMC gripper ($g_t$) is changed with a step size of $\sigma_h$ to close the AST finger assembly. Therefore, the grip width command at each time step reads as follows:
\begin{equation}
\label{gripeq}
    g_{t+1} = 
    \begin{cases} 
    g_t - \sigma_h & \text{if } f_m < f_d - \epsilon, \\
    g_t & \text{if }  f_d - \epsilon < f_m < f_d + \epsilon, \\
    g_t + \sigma_h & f_m > f_d + \epsilon.
    \end{cases}
\end{equation}

The grasp width change's step size ($\sigma_h$) is tuned based on the target grip force to avoid overshooting in grip force changes or significant control lags. In our experimental trials, the hyper-parameters are defined as $\sigma_h = 1$ mm and $\epsilon = 0.1$ N.

\subsection{Experimental Trials}
\begin{figure}[tb!]   
\centering
    \begin{subfigure}[t]{0.35\textwidth}  
    \includegraphics[width=\textwidth]{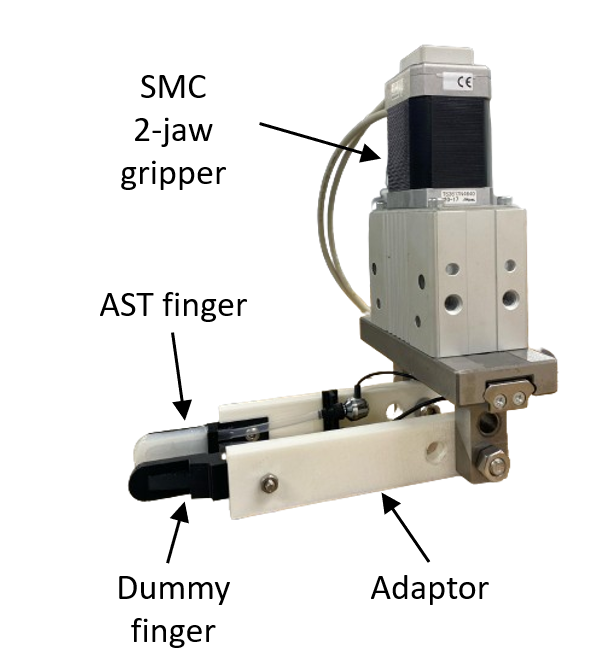}
    \caption{} \label{4a}
\end{subfigure}
\begin{subfigure}[t]{0.45\textwidth} 
    \includegraphics[width=\textwidth]{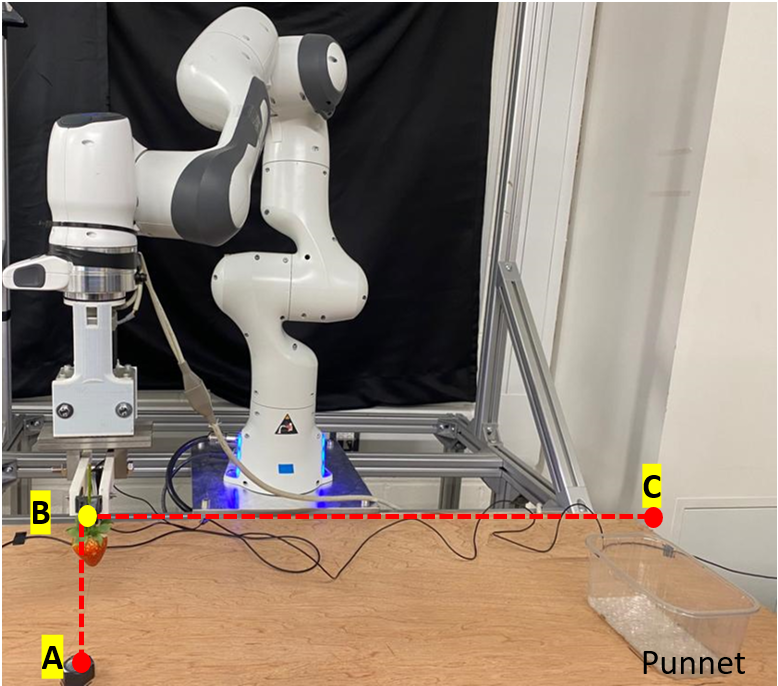}
    \caption{} \label{4b}
\end{subfigure}\hfill
\caption{ (a). AST finger assembly connected to the SMC gripper for gripping trials, (b). Experimental setup for the gripping trials: Point A is the picking point, B is a via point between A and C, and C is the dropping point}
\label{Fig1}
\end{figure}

The strawberries are placed individually in the picking location on the workbench with the peduncle upright (refer to figure~\ref{4b}).  At the start of the trial, the fingers will be positioned so that the peduncle is between the AST finger (calibrated area) and the dummy finger. Later, the end effector controller is triggered manually, and the fingers start closing with equal increments ($\sigma_h$ = 1 mm) until the gripping force value reaches the target gripping force ($f_d$=2N). Afterwards, the robot arm moves in a fixed trajectory towards the dropping point, where the strawberry is released into a punnet. The experiment scenario is shown in the figure~\ref{4b}.

This gripping trial is repeated five times for each strawberry sample, and the respective force readings from the AST finger ($f_m$) are logged in real-time to evaluate the performance. The upcoming section will discuss the analysis of the force readings.

\section{Results and Discussions}

\begin{figure}[tb!]
      \centering
        \includegraphics[width=0.6\textwidth]{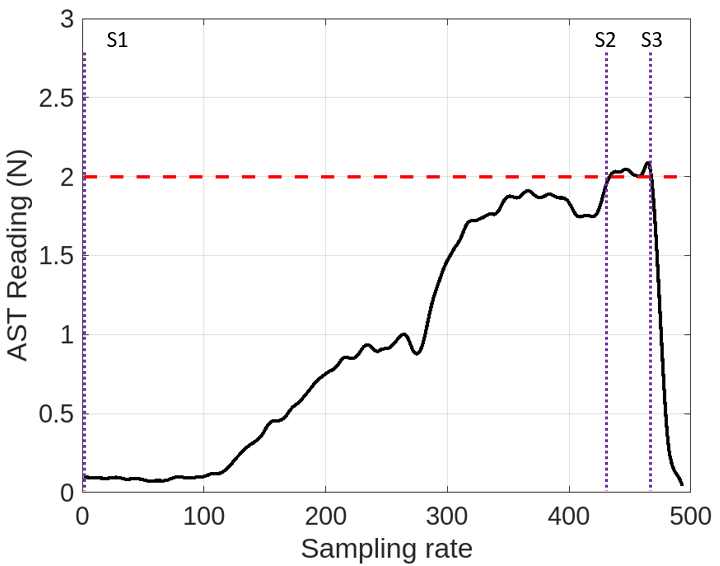}
\caption{AST readings ($f_m$) during a gripping sequence }
\label{senout}
\end{figure}

A sample of the force readings recorded during a gripping trial is shown in the figure~\ref{senout}. Three sections can be visualised from the force profile. In section S1, the peduncle comes in contact with the fingers. After S1, contact force increases as the fingers continue closing/gripping until the set limit ($f_d$) of 2 N is reached (in section S2). Later, the end effector travels from the picking to the dropping location (from point A to C via B), and force readings during this traverse are between S2 and S3 in figure~\ref{senout}. In section S3, the grip force suddenly drops due to the opening of fingers to release the strawberry into the punnet.

From this force profile, the readings ($f_m$) between S2 and S3 signify how close the grip force settles around the desired grip force ($f_d$= 2N). The table~\ref{MAE} presents the Mean Absolute Error (MAE), which quantifies the variation of $f_m$ from $f_d$. It has been studied that the maximum MAE is about 0.31 N, which is negligible for this task as it doesn't damage any of the peduncles. It is to be noted that this variation in sensor reading can be for two reasons: (i). the effect of swinging the strawberry during the traverse from picking to dropping location, or (ii). the inherent calibration error of the AST skin, as discussed in section 2.3. 

The results of the above study provide confidence in the usability of AST skin-embedded fingers in force-controlled gripping tasks using its feedback. 

\begin{table}[]
\centering
\caption{Mean Absolute Error (MAE) recorded for each gripping trial against the set grip force value of 2 N}
\label{MAE}
\begin{tabular}{|c|ccccc|c|}
\hline
\multirow{2}{*}{\textbf{Sample}} & \multicolumn{5}{c|}{\textbf{Trials}}                                                                                                               & \multirow{2}{*}{\textbf{\begin{tabular}[c]{@{}c@{}}Average \\  MAE\end{tabular}}} \\ \cline{2-6}
                                 & \multicolumn{1}{c|}{\textbf{1}} & \multicolumn{1}{c|}{\textbf{2}} & \multicolumn{1}{c|}{\textbf{3}} & \multicolumn{1}{c|}{\textbf{4}} & \textbf{5} &                                                                                   \\ \hline
1                                & \multicolumn{1}{c|}{0.079}      & \multicolumn{1}{c|}{0.167}      & \multicolumn{1}{c|}{0.143}      & \multicolumn{1}{c|}{0.102}      & 0.089      & 0.116                                                                             \\ \hline
2                                & \multicolumn{1}{c|}{0.082}      & \multicolumn{1}{c|}{0.684}      & \multicolumn{1}{c|}{0.560}      & \multicolumn{1}{c|}{0.047}      & 0.182      & 0.311                                                                             \\ \hline
3                                & \multicolumn{1}{c|}{0.219}      & \multicolumn{1}{c|}{0.307}      & \multicolumn{1}{c|}{0.401}      & \multicolumn{1}{c|}{0.178}      & 0.358      & 0.293                                                                             \\ \hline
4                                & \multicolumn{1}{c|}{0.238}      & \multicolumn{1}{c|}{0.202}      & \multicolumn{1}{c|}{0.095}      & \multicolumn{1}{c|}{0.050}      & 0.222      & 0.161                                                                             \\ \hline
5                                & \multicolumn{1}{c|}{0.106}      & \multicolumn{1}{c|}{0.059}      & \multicolumn{1}{c|}{0.142}      & \multicolumn{1}{c|}{0.031}      & 0.115      & 0.090                                                                             \\ \hline
\end{tabular}
\end{table}

\section{Conclusion}

This paper tests the usability of Acoustic Soft Tactile (AST) skin as a customisable tactile skin for enabling tactile feedback for robot manipulation tasks. This has been evaluated by building and testing an end effector finger with AST skin to perform force-controlled strawberry gripping. The AST finger facilitated the end effector to handle the strawberry by gripping its peduncle with a set target force of 2 N with a maximum mean absolute error of 0.31 N. 
As a future work, the AST fingers will be integrated into the custom-built end effector and tested in the field by mounting it on a strawberry harvesting robot~\cite{parsa2023modular}. Moreover, the entire skin area will be calibrated to measure the gripping force with the description of the peduncle location on the skin surface. This contact localisation approach helps confirming a successful grip at the target locked by the robot's integrated vision system.

\bibliographystyle{splncs04}  
\bibliography{Reference}  
\end{document}